\documentclass[conference]{IEEEtran}
\IEEEoverridecommandlockouts
\usepackage[noend]{algpseudocode}
\usepackage{amsmath,amssymb,amsfonts}
\usepackage{booktabs}
\usepackage{algorithm}
\usepackage{graphicx}
\usepackage{subcaption}
\usepackage{textcomp}
\usepackage[table]{xcolor}
\usepackage{url}
\usepackage{array}
\usepackage{arydshln}
\usepackage{float}
\usepackage{hyperref}
\usepackage{multirow}
\def\BibTeX{{\rm B\kern-.05em{\sc i\kern-.025em b}\kern-.08em
    T\kern-.1667em\lower.7ex\hbox{E}\kern-.125emX}}
\begin{document}

\title{Scalpel-SAM: A Semi-Supervised Paradigm for Adapting SAM to Infrared Small Object Detection}

\author{
    \IEEEauthorblockN{Zihan Liu\IEEEauthorrefmark{1}, Xiangning Ren\IEEEauthorrefmark{2}, Dezhang Kong\IEEEauthorrefmark{3}, Yipeng Zhang\IEEEauthorrefmark{4}, and Meng Han\IEEEauthorrefmark{3}}
    \IEEEauthorblockA{\IEEEauthorrefmark{1}School of Computer Science, Chengdu University, Chengdu, 610106, China}
    \IEEEauthorblockA{\IEEEauthorrefmark{2}China University of Geosciences, Beijing, China}
    \IEEEauthorblockA{\IEEEauthorrefmark{3}Zhejiang University, Hangzhou, China}
    \IEEEauthorblockA{\IEEEauthorrefmark{4}Intelligent Computing Infrastructure Innovation Center, Zhejiang Lab, Hangzhou, China}
}

\maketitle

\begin{abstract}
Infrared small object detection urgently requires semi-supervised paradigms due to the high cost of annotation. However, existing methods like SAM face significant challenges of domain gaps, inability of encoding physical priors, and inherent architectural complexity. To address this, we designed a Hierarchical MoE Adapter consisting of four white-box neural operators. Building upon this core component, we propose a two-stage paradigm for knowledge distillation and transfer: (1) Prior-Guided Knowledge Distillation, where we use our MoE adapter and 10\% of available fully supervised data to distill SAM into an expert teacher (Scalpel-SAM); and (2) Deployment-Oriented Knowledge Transfer, where we use Scalpel-SAM to generate pseudo labels for training lightweight and efficient downstream models. Experiments demonstrate that with minimal annotations, our paradigm enables downstream models to achieve performance comparable to, or even surpassing, their fully supervised counterparts. To our knowledge, this is the first semi-supervised paradigm that systematically addresses the data scarcity issue in IR-SOT using SAM as the teacher model.
\end{abstract}

\begin{IEEEkeywords}
Semi-Supervised Learning, Infrared Small Object Detection, Segment Anything Model
\end{IEEEkeywords}

\section{Introduction}
Infrared small object detection (IR-SOT) serves as a foundational technology in defense security, disaster rescue, and other critical areas, owing to its unique capability to identify thermal signatures under all-weather conditions and in total darkness where visible light systems fail \cite{zhao2022single, kou2023survey}. However, IR-SOT is being plagued by two major bottlenecks: First, infrared images have extremely minute targets, low signal-to-noise ratios, and a severe lack of textural cues and color cues that are abundant in natural images\cite{zhao2022single}. Second, obtaining pixel-level ground truth annotations incurs an exceedingly high cost \cite{kou2023survey}. The laborious and challenging process of manual labeling makes traditional methods, which rely on large-scale supervision, impractical. As such, exploring how to achieve efficient detection in the face of scarce data has become an urgent demand in this field.

\begin{figure}[t]
    \centering
    \begin{subfigure}[b]{0.5\textwidth} 
        \centering
        \includegraphics[width=\textwidth]{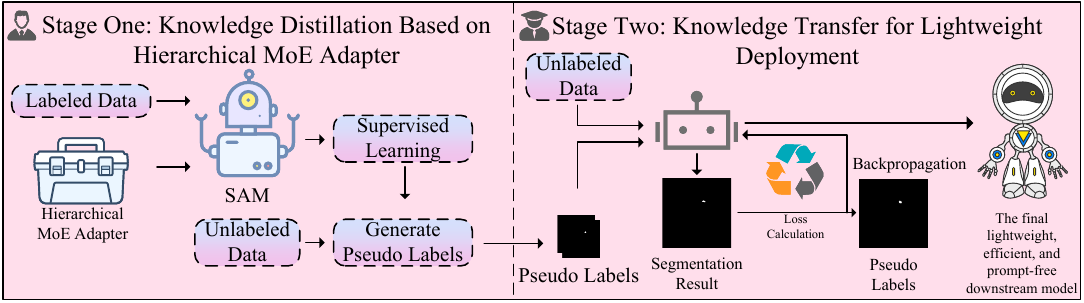}
        \caption{}
        \label{fig:sub_a}
    \end{subfigure}
    \\
    \begin{subfigure}[b]{0.5\textwidth}
        \centering
        \includegraphics[width=\textwidth]{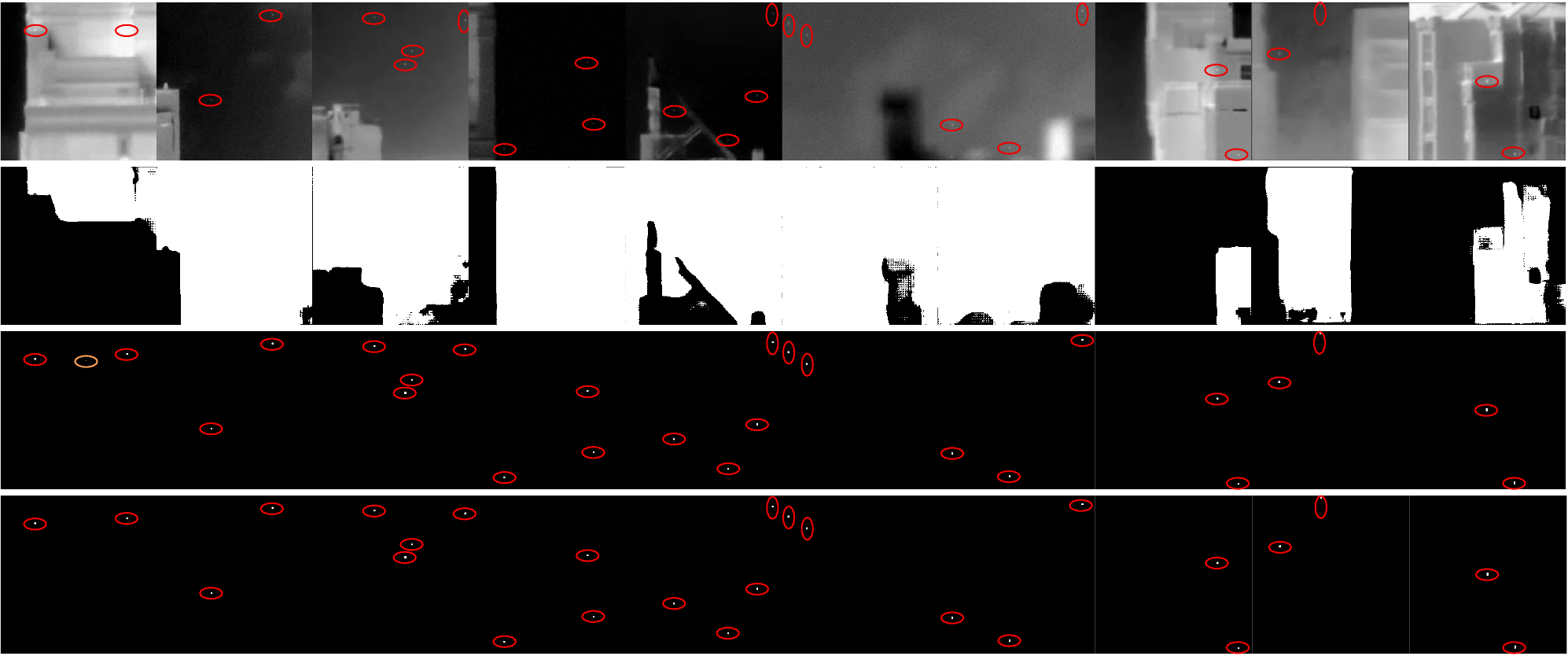}
        \caption{}
        \label{fig:sub_b}
    \end{subfigure}
    \caption{The proposed paradigm and the visual evidence of the challenge.
(a) Our two-stage paradigm. Stage One refines SAM (robot icon) into an expert teacher via our MoE adapter; Stage Two trains a lightweight student model (white robot) using teacher-generated pseudo-labels.
(b) Visual comparison on the SIRST3 dataset. From top to bottom, the rows display: the original image, the generic SAM's segmentation, our Scalpel-SAM's precise result, and the GT.}
    \label{fig:Introduction}
\end{figure}

Large foundation models, exemplified by Segment Anything Model (SAM)\cite{sam}, offer a solution to this impasse. This is because SAM's unique prompt-driven architecture can handle minimal, sparse guidance and output high-quality, dense segmentation masks. 
However, SAM was not designed for the specific physical properties of IR-SOT. Therefore, when we attempt to directly apply it to this specialized task, there are three challenges that must be addressed: (1) severe performance failure due to the domain gap, as visually demonstrated in Figure \ref{fig:sub_b}, where generic SAM (second row) produces noisy and failed segmentations; (2) the black-box limitations of Parameter-Efficient Fine-Tuning (PEFT) which prevent encoding physical priors; and (3) the inherent architectural complexity of SAM.
To systematically address these three challenges, we propose a two-stage knowledge distillation and transfer paradigm, as shown in Figure \ref{fig:sub_a}. The Stage One addresses challenges 1 and 2, where we design an innovative Hierarchical MoE Adapter to fine-tune SAM into an expert teacher called Scalpel-SAM. Its core is an expert toolkit composed of four white-box neural operators (Physics-Informed Manifold Diffusion Operator (PIMDO), Spectral Prism Decoupler (SPD), Hyper-Parameterized Local Statistics Modulator (HPLSM), Topology-Guided Deformable Sampler (TGDS)). These operators explicitly encode anisotropic diffusion, frequency decoupling, local statistical adaptation, and topological connectivity priors into the network structure, elevating PEFT from a generic black-box into an interpretable regulator with surgical precision. The Stage Two tackles challenge 3, where we use the expert teacher trained in the first stage and generate high-quality pseudo labels. This pivotal step both amplifies the knowledge learned from the 10\% supervised data and enables the knowledge to be transferred to a lightweight, efficient, and prompt-free student model, thereby drastically reducing the final deployment cost. This paradigm successfully transfers SAM’s powerful prior knowledge into a deployable lightweight model while solving the dual dilemmas of model capability and data costs. 

Extensive experiments validate the efficacy of our paradigm. The results demonstrate that by applying our paradigm, the performance of downstream models approaches or \textbf{even surpasses that of their fully supervised counterparts}. To our knowledge, no prior semi-supervised paradigm in IR-SOT has managed to close this performance gap, let alone exceed the fully supervised upper bound.
We also publish our code on \textbf{\url{https://anonymous.4open.science/r/Scalpel-SAM}(Anonymized for review)}. The contributions of this paper can be summarized as follows:
\begin{itemize}
\item We propose a two-stage knowledge distillation and transfer paradigm, which can automatically label vast data and help train downstream models. To our knowledge, this is the first semi-supervised learning paradigm for IR-SOT detection.
\item We design a Hierarchical MoE Adapter that replaces the generic black-box structure of traditional PEFT methods with a domain-principle-driven expert toolkit, which can mimic a human expert's progressive thought chain for target analysis and elevate PEFT into a domain-driven, highly interpretable adaptive regulator.
\item Comprehensive experiments demonstrate that our paradigm approaches or even surpasses fully supervised performance using only 10\% of the labeled data.
\end{itemize}
\begin{table}[htbp]
  \centering
  \tiny
  \caption{Symbols and Notations}
  \label{tab:notation}
  \begin{tabular}{lll}
    \hline
    \textbf{Symbol} & \textbf{Description} \\
    \hline
    \multicolumn{2}{l}{\textbf{Models, Data, and General Parameters}} \\
    \hline
    $D_{L}, D_{U}$ & Supervised (labeled) and unsupervised (unlabeled) datasets \\
    $D_{pseudo}$ & Generated pseudo label dataset \\
    $\mathcal{E}_{\Phi}$ & Frozen SAM ViT Encoder backbone \\
    $\Phi$ & Frozen parameters of $\mathcal{E}_{\Phi}$ \\
    $\mathcal{A}_{\theta}$ & Trainable MoE Adapter with parameters $\theta$ \\
    $\theta$ & Trainable parameters of $\mathcal{A}_{\theta}$ (in Stage One) \\
    $\theta^{*}$ & Optimized parameters of $\mathcal{A}_{\theta}$ after Stage One training \\
    $\mathcal{A}_{\theta^{*}}$ & Expert teacher adapter with optimized parameters $\theta^{*}$ \\
    $\mathcal{S}_{\psi}$ & Lightweight downstream student model \\
    $\psi$ & Trainable parameters of $\mathcal{S}_{\psi}$ (in Stage Two) \\
    $I, M^{gt}$ & Input infrared image and its Ground Truth label \\
    $M^{pseudo}$ & Pseudo label generated by the teacher model \\
    $M_{pred}$ & Segmentation prediction output (from teacher or student model) \\
    \hline
    \multicolumn{3}{l}{\textbf{Core Component: MoE Adapter}} \\
    \hline
    $\mathcal{B}_l$ & $l$-th Transformer module \\
    $x_l$ & Input features to the $l$-th layer \\
    $\mathcal{A}^{(l)}$ & Adapter module injected into the $l$-th layer \\
    $\mathcal{R}(\cdot)$ & Dynamic routing network of the MoE \\
    $\mathcal{E}_{i}(\cdot)$ & $i$-th white-box expert operator \\
    $K$ & Total number of experts (K=4) \\
    $w_i$ & Weight assigned to the $i$-th expert by the router \\
    $c(\cdot)$ & Learnable intelligent diffusion controller in PIMDO \\
    $\mathcal{W}_{ana}, \mathcal{W}_{syn}$ & Learnable wavelet analysis and synthesis transforms in SPD \\
    $\mathcal{M}_{spec}(\cdot)$ & Spectral attention module in SPD \\
    $\mathcal{H}_{\psi}(\cdot)$ & Hypernetwork in HPLSM \\
    $(\gamma_{ij}, \beta_{ij})$ & Dynamically generated affine parameters by HPLSM \\
    $\mathcal{G}_{off}(\cdot)$ & Offset field prediction network in TGDS \\
    \hline
    \multicolumn{2}{l}{\textbf{Loss Functions}} \\
    \hline
    $\mathcal{L}_{total}$ & Total loss for Stage One \\
    $\mathcal{L}_{main}$ & Main segmentation task loss (BCE + Dice) \\
    $\mathcal{L}_{BCE}, \mathcal{L}_{Dice}$ & Binary Cross-Entropy loss and Dice loss \\
    $\mathcal{L}_{sparse}$ & Auxiliary sparsity loss for MoE  \\
    $\mathcal{L}_{topo}$ & Topological regularization loss in TGDS \\
    $\lambda_{bce}, \lambda_{dice}$ & Weights for BCE and Dice losses \\
    $\lambda_{topo}$ & Weight for the topological loss \\
    $\lambda_{sparse}$ & Weight for the sparsity loss \\
    $f_i, P_i$ & Selection frequency and average probability of expert $i$ in $\mathcal{L}_{sparse}$ \\
    \hline
  \end{tabular}
\end{table}

\section{Methodology}
For clarity, all mathematical symbols and notations used in our methodology are shown in Table \ref{tab:notation}.
\subsection{Overview: Knowledge Distillation and Transfer Paradigm}
\label{sec:framework}
\begin{figure}[t]
    \centering
    \includegraphics[width=0.46\textwidth]{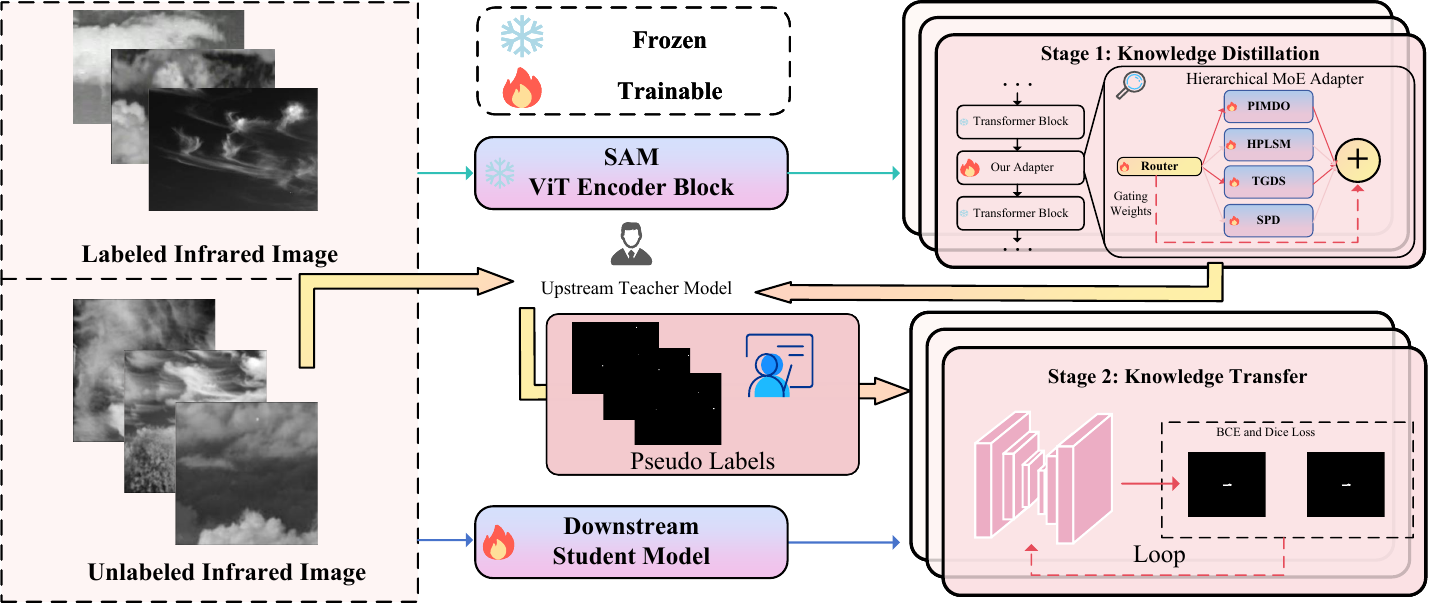}
    \caption{The architecture of our two-stage paradigm. Stage 1 (Top-Right): Our trainable Hierarchical MoE Adapter (containing a router and four expert operators: PIMDO, SPD, HPLSM, TGDS) is injected into the frozen SAM ViT Encoder. Stage 2 (Bottom-Right): The resulting expert teacher generates pseudo labels from unlabeled data to supervise a lightweight downstream student.}
    \label{fig:Total_Process}
\end{figure}
Our paradigm decouples the complex adaptation and learning process into two distinct stages.

\subsubsection{Stage One: Knowledge Distillation Based on Hierarchical MoE Adapter}
The goal of this stage is to utilize a minimal amount of all available fully supervised data $D_{L}$ to refine the general SAM into an expert pseudo label generator Scalpel-SAM. To achieve this, we inject a Hierarchical MoE Adapter $\mathcal{A}_{\theta}$ into the last two Transformer modules following the ViT encoder $\mathcal{E}_{\Phi}$ of SAM. During fine-tuning, only the adapter parameters $\theta$ are updated, while the vast backbone network parameters $\Phi$ of SAM remain frozen. This process can be formalized as:
\begin{equation}
\resizebox{0.9\linewidth}{!}{
    $\theta = \arg\min_{\theta} \mathcal{L}_{task}(\mathcal{E}_{\Phi}(I; \mathcal{A}_{\theta}), M^{gt}), \quad \forall (I, M^{gt}) \in D_L
    $
    }
\end{equation}

\subsubsection{Stage Two: Knowledge Transfer for Lightweight Deployment}
The goal of this stage is to distill knowledge from the expert teacher into an efficient, deployable, and prompt-free downstream model $\mathcal{S}_{\psi}$. We denote the optimized expert adapter from Stage One as $\mathcal{A}_{\theta^{*}}$. This adapter is first used to generate pseudo-masks ($M_j^{pseudo}$) for the entire training set $D_{U}$. This yields a complete pseudo label dataset, $D_{pseudo}$, which then serves as the sole supervision to train the lightweight network $\mathcal{S}_{\psi}$ from scratch.

\subsection{Stage One: Knowledge Distillation Based on Hierarchical MoE Adapter}
\label{sec:adapter}
The complete training procedure of Stage One is detailed in Algorithm \ref{alg:stage_one}, and its architecture is illustrated in Figure \ref{fig:Total_Process}.
\begin{algorithm}[h!]
\footnotesize
\caption{Stage 1: Knowledge Distillation via Hierarchical MoE Adapter}
\label{alg:stage_one}
\begin{algorithmic}[1]
    \Require Frozen SAM ViT Encoder $\mathcal{E}_{\Phi}$, Trainable MoE Adapter $\mathcal{A}_{\theta}$ (with Router $\mathcal{R}_{\theta}$ and Experts $\mathcal{E}_{\theta}^{(i)}$), Small fully supervised dataset $D_{L} = \{(I, M^{gt})\}$, Sparsity weight $\lambda_{sparse}$.
    \Ensure Trained adapter parameters $\theta^{*}$.
    \vspace{-0.5mm}
    \State $\theta \gets \text{InitializeAdapterParameters}()$
    \For{each epoch}
        \For{each $(I, M^{gt}) \in D_{L}$}
            \State $x \gets \text{PatchEmbed}(I)$ 
            
            \For{$l = 1$ to $L$} 
                \State $x_{block\_out} \gets \mathcal{B}_l(x)$ 
                \If{$l > L - 2$}
                    \State $x_{in} \gets x$ 
                    
                    \State $w \gets \mathcal{R}_{\theta}(x_{in})$ \text{(See Section \ref{ssub:router})} 
                    
                    \State $\mathcal{A}^{(l)}(x_{in}) \gets 0$
                    \For{$i = 1$ to $K=4$} 
                        \State $\mathcal{A}^{(l)}(x_{in}) \gets \mathcal{A}^{(l)}(x_{in}) + w_i \cdot \mathcal{E}_{\theta}^{(i)}(x_{in})$ \text{(See Section \ref{ssub:router} \& \ref{ssub:toolbox})}
                    \EndFor
                    
                    \State $x \gets x_{block\_out} + \mathcal{A}^{(l)}(x_{in})$ (See Section \ref{ssub:toolbox})
                \Else
                    \State $x \gets x_{block\_out}$
                \EndIf
            \EndFor
            
            \State $M_{pred} \gets \text{Decoder}(x)$ 
            
            \State $\mathcal{L}_{main} \gets \mathcal{L}_{BCE}(M_{pred}, M^{gt}) + $
            \Statex \hspace{\algorithmicindent} $\mathcal{L}_{Dice}(M_{pred}, M^{gt})$ \text{(See Section \ref{ssub:stage_one_training_objectives})}
            \State $\mathcal{L}_{sparse} \gets \text{LoadBalancingLoss}(w)$ 
            \Statex \hspace{\algorithmicindent} \text{(See Section \ref{ssub:stage_one_training_objectives})}
            \State $\mathcal{L}_{total} \gets \mathcal{L}_{main} + \lambda_{sparse} \cdot \mathcal{L}_{sparse}$ 
            \Statex \hspace{\algorithmicindent} \text{(See Section \ref{ssub:stage_one_training_objectives})}
            
            \State $\text{Update}(\theta)$ 
        \EndFor
    \EndFor
    \State \textbf{return} $\theta^{*}$
\end{algorithmic}
\end{algorithm}
\subsubsection{Dynamic Expert Routing Mechanism}
\label{ssub:router}
We employ a lightweight routing network $\mathcal{R}(\cdot)$ to dynamically compute weights $w_i$ for our $K$ parallel experts. We adopt a soft selection mechanism, where the final adapter output is a weighted fusion of all expert outputs $\mathcal{E}_i(x_l)$. Full formulas are provided in the supplementary material \S\ref{sec:supp_adapter}.

\subsubsection{White-Box Expert Toolbox}
\label{ssub:toolbox}

The four neural operators we designed collectively mimic the progressive cognitive chain by which human experts analyze infrared images: (1) Signal Refining, separating suspicious signals from complex backgrounds (PIMDO \& SPD); (2) Contextual Confirmation, determining signal authenticity based on local context (HPLSM); and (3) Morphological Focusing, delineating the target shape (TGDS) .

\noindent $\bullet$ \textbf{Physics-Informed Manifold Diffusion Operator (PIMDO)}
To mimic the first step of this chain (refining signals), we propose the PIMDO operator, which we formulate as a learnable anisotropic diffusion process. Its continuous form is governed by the PDE:
\begin{equation}
\frac{\partial x}{\partial t} = \nabla \cdot (c(|\nabla x|) \nabla x)
\label{eq:pimdo}
\end{equation}
The core of PIMDO is an intelligent diffusion controller $c(\cdot)$, which is trained to promote smoothing in regions with gentle gradients (i.e., noise) while suppressing diffusion at the steep gradients of target edges. However, operating solely in the spatial domain is insufficient. In IR-SOT tasks, many high-frequency background clutter exhibit spatial features highly similar to real targets.

\noindent $\bullet$ \textbf{Spectral Prism Decoupler (SPD)}
We propose the SPD operator, which aims to decouple the feature map into orthogonal subspaces $\{z_j\}$, representing different scales and frequency information, using a learnable wavelet-like transform $\mathcal{W}_{ana}$. Then, a spectral attention module $\mathcal{M}_{spec}(\cdot)$ automatically identifies and enhances the subspaces most likely to contain the target signal while suppressing those that contain clutter. Finally, the feature map is reconstructed through the inverse transform $\mathcal{W}_{syn}$:
\begin{equation}
    x'_{l} = \mathcal{W}_{syn}(\{\alpha_j z_j\}_{j=1}^{J}), \quad \text{where } \alpha = \mathcal{M}_{spec}(\{z_j\})
    \label{eq:spd}
\end{equation}
However, static decoupling fails in the highly dynamic and non-homogeneous backgrounds of infrared images (e.g., sky vs. ground clutter), which necessitates the adaptive mechanism introduced next.

\noindent $\bullet$ \textbf{Hyper-Parameterized Local Statistics Modulator (HPLSM)}
This operator endows standard convolutions with environmental awareness capabilities. It contains a hypernetwork $\mathcal{H}_{\psi}(\cdot)$, which analyzes the local statistics of the input feature $x_l$ in real-time and dynamically generates an affine transformation parameter pair $(\gamma_{ij}, \beta_{ij})$ for each spatial location $(i,j)$. These parameters then modulate the output of a shared base convolution kernel $\text{Conv}_{\theta}$:
\begin{equation}
    \text{Output}(x_l)_{ij} = \gamma_{ij} \odot \text{Conv}_{\theta}(x_l)_{ij} + \beta_{ij}
    \label{eq:hplsm}
\end{equation}
At this point, the target signal has been largely refined, but precise localization in the final step remains a challenge.

\noindent $\bullet$ \textbf{Topology-Guided Deformable Sampler (TGDS)}
We design the TGDS, which employs an offset field prediction network $\mathcal{G}_{off}(\cdot)$, allowing deformable convolutions to actively and non-grid-like capture the target. More importantly, we introduce a crucial topological regularization term $\mathcal{L}_{topo}$, which encourages the sampled points ${p_k + \Delta p_k}$ to form a continuous structure that aligns with the target's geometric shape. The overall offset learning objective is:
\begin{equation}
    \mathcal{L}_{offset} = \mathcal{L}_{task} + \lambda_{topo} \cdot \mathcal{L}_{topo}({p_k + \Delta p_k})
    \label{eq:tgds}
\end{equation}

\subsubsection{Training Objectives}
\label{ssub:stage_one_training_objectives}
The total loss $\mathcal{L}_{total}$ consists of a main task loss $\mathcal{L}_{main}$ and an auxiliary sparsity loss $\mathcal{L}_{sparse}$. The main loss is a standard weighted sum of BCE and Dice:
\begin{equation}
\resizebox{.9\hsize}{!}{$
\mathcal{L}_{main} = \lambda_{bce} \cdot \mathcal{L}_{BCE}(M_{pred}, M_{gt}) + \lambda_{dice} \cdot \mathcal{L}_{Dice}(M_{pred}, M_{gt})
$}
\end{equation}
The sparsity loss $\mathcal{L}_{sparse}$ is a standard load balancing loss to encourage expert specialization:
\begin{equation}
\mathcal{L}_{sparse} = \alpha \cdot K \sum_{i=1}^{K} f_i \cdot P_i
\end{equation}
where $f_i$ and $P_i$ are the selection frequency and average probability for expert $i$, respectively. The final weighted loss is:
\begin{equation}
\mathcal{L}_{total} = \mathcal{L}_{main} + \lambda_{sparse} \cdot \mathcal{L}_{sparse}
\end{equation}

\begin{table*}[htbp]
    \centering
    \tiny
    \caption{Performance comparison. \textcolor[HTML]{056f28}{\textbf{Green numbers}} represent our paradigm's results that exceed the SOTA point-supervised paradigm. For those results that even surpass the fully supervised paradigm, we use \textcolor{red}{\textbf{red numbers}} to highlight them. }
    \label{tab:comparison}
    \resizebox{\textwidth}{!}{
    \begin{tabular}{llcccccccccccccccc}
        \hline
        \multirow{2}{*}{Scheme} & \multirow{2}{*}{Description} & \multicolumn{4}{c}{SIRST3-Test} & \multicolumn{4}{c}{NUAA-SIRST-Test} & \multicolumn{4}{c}{NUDT-SIRST-Test} & \multicolumn{4}{c}{IRSTD-1K-Test} \\
        \cline{3-18}
         & & $mIoU$ & $nIoU$ & $P_d$ & $F_a$ & $mIoU$ & $nIoU$ & $P_d$ & $F_a$ & $mIoU$ & $nIoU$ & $P_d$ & $F_a$ & $mIoU$ & $nIoU$ & $P_d$ & $F_a$ \\
        \hline
        \multirow{3}{*}{ACM\cite{NUAA-SIRST}}& DLN Full & 62.52 & 62.36 & 93.09 & 32.27 & 67.36 & 66.25 & 92.02 & 28.881 & 68.89 & 69.61 & 97.25 & 11.283 & 63.05 & 57.06 & 89.23 & 30.233 \\
         & DLN Course + PAL & 50.4 & 51.71 & 92.29 & 30.338 & 55.33 & 56.61 & 85.67 & 27.89 & 53.83 & 52.67 & 92.60 & 18.31 & 30.58 & 41.76 & 78.45 & 18.295 \\
         & Ours & \textcolor[HTML]{056f28}{\textbf{56.9 (91\%)}} & \textcolor[HTML]{056f28}{\textbf{54.37}} & \textcolor[HTML]{056f28}{\textbf{92.89}} & \textcolor[HTML]{056f28}{\textbf{28.641}} & \textcolor[HTML]{056f28}{\textbf{61.81 (92\%)}} & \textcolor[HTML]{056f28}{\textbf{59.69}} & \textcolor[HTML]{056f28}{\textbf{87.45}} & \textcolor[HTML]{056f28}{\textbf{32.791}} & \textcolor[HTML]{056f28}{\textbf{60.23 (87\%)}} & \textcolor[HTML]{056f28}{\textbf{61.42}} & \textcolor[HTML]{056f28}{\textbf{93.65}} & \textcolor[HTML]{056f28}{\textbf{21.578}} & \textcolor{red}{\textbf{65.26 (104\%)}} & \textcolor[HTML]{056f28}{\textbf{59.23}} & \textcolor[HTML]{056f28}{\textbf{92.26}} & \textcolor[HTML]{056f28}{\textbf{36.705}} \\
        \hdashline
        \multirow{3}{*}{ALCNet\cite{ALCNet}} & DLN Full & 64.35 & 66.22 & 93.02 & 30.97 & 69.25 & 67.49 & 92.45 & 14.82 & 64.96 & 63.26 & 96.61 & 12.547 & 62.78 & 56.47 & 90.57 & 36.135 \\
         & DLN Course + PAL & 43.91 & 48.44 & 87.51 & 18.243 & 27.99 & 31.06 & 70.72 & 8.095 & 55.63 & 54.80 & 90.20 & 39.83 & 12.52 & 15.63 & 34.34 & 1.537 \\
         & Ours & \textcolor[HTML]{056f28}{\textbf{61.78 (96\%)}} & \textcolor[HTML]{056f28}{\textbf{64.1}} & \textcolor[HTML]{056f28}{\textbf{96.01}} & \textcolor[HTML]{056f28}{\textbf{49.457}} & \textcolor[HTML]{056f28}{\textbf{52.02 (75\%)}} & \textcolor[HTML]{056f28}{\textbf{53.01}} & \textcolor[HTML]{056f28}{\textbf{83.65}} & \textcolor[HTML]{056f28}{\textbf{56.877}} & \textcolor[HTML]{056f28}{\textbf{63.51 (98\%)}} & \textcolor[HTML]{056f28}{\textbf{62.62}} & \textcolor[HTML]{056f28}{\textbf{95.51}} & \textcolor[HTML]{056f28}{\textbf{15.673}} & \textcolor[HTML]{056f28}{\textbf{61.38 (98\%)}} & \textcolor[HTML]{056f28}{\textbf{55.53}} & \textcolor[HTML]{056f28}{\textbf{90.36}} & \textcolor[HTML]{056f28}{\textbf{15.680}} \\
        \hdashline
        \multirow{3}{*}{MLCL-Net\cite{MLCLNet}} & DLN Full & 79.79 & 82.97 & 95.81 & 15.598 & 70.88 & 73.25 & 93.92 & 44.728 & 94.36 & 93.96 & 99.05 & 2.689 & 65.12 & 64.42 & 91.92 & 15.619 \\
         & DLN Course + PAL & 64.28 & 69.55 & 94.95 & 24.652 & 67.35 & 69.67 & 94.30 & 48.638 & 73.38 & 74.3 & 96.72 & 19.993 & 59.89 & 58.32 & 90.24 & 18.922 \\
         & Ours & \textcolor[HTML]{056f28}{\textbf{71.24 (89\%)}} & \textcolor[HTML]{056f28}{\textbf{75.00}} & \textcolor[HTML]{056f28}{\textbf{95.95}} & \textcolor[HTML]{056f28}{\textbf{32.839}} & \textcolor{red}{\textbf{72.09 (102\%)}} & \textcolor[HTML]{056f28}{\textbf{72.74}} & \textcolor[HTML]{056f28}{\textbf{95.06}} & \textcolor[HTML]{056f28}{\textbf{28.744}} & \textcolor[HTML]{056f28}{\textbf{75.56 (80\%)}} & \textcolor[HTML]{056f28}{\textbf{76.85}} & \textcolor[HTML]{056f28}{\textbf{98.08}} & \textcolor[HTML]{056f28}{\textbf{15.326}} & \textcolor[HTML]{056f28}{\textbf{64.39 (99\%)}} & \textcolor[HTML]{056f28}{\textbf{63.31}} & \textcolor[HTML]{056f28}{\textbf{92.26}} & \textcolor[HTML]{056f28}{\textbf{18.713}} \\
        \hdashline
        \multirow{3}{*}{ALCL-Net\cite{ALCLNet}} & DLN Full & 69.22 & 70.59 & 93.49 & 36.973 & 64.22 & 65.24 & 93.92 & 67.572 & 65.54 & 62.73 & 90.91 & 49.515 & 66.46 & 62.53 & 87.21 & 41.525 \\
         & DLN Course + PAL & 63.61 & 67.97 & 95.15 & 21.438 & 57.21 & 57.6 & 85.17 & 30.322 & 68.84 & 70.84 & 98.41 & 11.996 & 51.52 & 56.85 & 89.56 & 27.842 \\
         & Ours & \textcolor{red}{\textbf{70.84 (102\%)}} & \textcolor[HTML]{056f28}{\textbf{73.92}} & \textcolor[HTML]{056f28}{\textbf{94.68}} & \textcolor[HTML]{056f28}{\textbf{21.402}} & \textcolor[HTML]{056f28}{\textbf{59.88 (93\%)}} & \textcolor[HTML]{056f28}{\textbf{59.74}} & \textcolor[HTML]{056f28}{\textbf{91.63}} & \textcolor[HTML]{056f28}{\textbf{48.775}} & \textcolor{red}{\textbf{71.08 (108\%)}} & \textcolor[HTML]{056f28}{\textbf{73.57}} & \textcolor[HTML]{056f28}{\textbf{95.77}} & \textcolor[HTML]{056f28}{\textbf{26.427}} & \textcolor[HTML]{056f28}{\textbf{64.15 (97\%)}} & \textcolor[HTML]{056f28}{\textbf{62.47}} & \textcolor[HTML]{056f28}{\textbf{91.58}} & \textcolor[HTML]{056f28}{\textbf{19.947}} \\
        \hdashline
        \multirow{3}{*}{DNANet\cite{NUDT-SIRST}} & DLN Full & 81.56 & 85.12 & 96.15 & 10.417 & 69.39 & 72.6 & 94.68 & 62.084 & 85.95 & 87.00 & 98.52 & 9.973 & 63.75 & 63.06 & 90.24 & 19.472 \\
         & DLN Course + PAL & 64.31 & 70.33 & 95.02 & 12.366 & 64.96 & 66.98 & 89.73 & 25.108 & 67.75 & 66.66 & 97.67 & 6.71 & 47.82 & 50.46 & 81.48 & 29.777 \\
         & Ours & \textcolor[HTML]{056f28}{\textbf{73.34 (90\%)}} & \textcolor[HTML]{056f28}{\textbf{76.98}} & \textcolor[HTML]{056f28}{\textbf{95.95}} & \textcolor[HTML]{056f28}{\textbf{27.531}} & \textcolor{red}{\textbf{73.82 (106\%)}} & \textcolor[HTML]{056f28}{\textbf{73.26}} & \textcolor[HTML]{056f28}{\textbf{93.54}} & \textcolor[HTML]{056f28}{\textbf{17.15}} & \textcolor[HTML]{056f28}{\textbf{72.68 (85\%)}} & \textcolor[HTML]{056f28}{\textbf{76.06}} & \textcolor[HTML]{056f28}{\textbf{97.46}} & \textcolor[HTML]{056f28}{\textbf{26.289}} & \textcolor{red}{\textbf{68.03 (107\%)}} & \textcolor[HTML]{056f28}{\textbf{61.48}} & \textcolor[HTML]{056f28}{\textbf{87.88}} & \textcolor[HTML]{056f28}{\textbf{17.157}} \\
        \hdashline
        \multirow{3}{*}{GGL-Net\cite{GGLNet}} & DLN Full & 80.89 & 84.16 & 97.54 & 13.296 & 70.82 & 71.56 & 94.68 & 43.63 & 66.72 & 64.97 & 93.60 & 32.396 & 65.96 & 64.14 & 93.80 & 32.415 \\
         & DLN Course + PAL & 63.57 & 69.58 & 94.49 & 16.844 & 51.86 & 52.12 & 84.79 & 47.266 & 66.73 & 68.32 & 96.93 & 12.754 & 42.49 & 47.47 & 83.84 & 28.126 \\
         & Ours & \textcolor[HTML]{056f28}{\textbf{70.55 (87\%)}} & \textcolor[HTML]{056f28}{\textbf{73.84}} & \textcolor[HTML]{056f28}{\textbf{95.55}} & \textcolor[HTML]{056f28}{\textbf{28.091}} & \textcolor{red}{\textbf{72.48 (102\%)}} & \textcolor[HTML]{056f28}{\textbf{72.47}} & \textcolor[HTML]{056f28}{\textbf{95.44}} & \textcolor[HTML]{056f28}{\textbf{22.981}} & \textcolor{red}{\textbf{71.29 (107\%)}} & \textcolor[HTML]{056f28}{\textbf{73.68}} & \textcolor[HTML]{056f28}{\textbf{96.51}} & \textcolor[HTML]{056f28}{\textbf{27.369}} & \textcolor[HTML]{056f28}{\textbf{64.03 (97\%)}} & \textcolor[HTML]{056f28}{\textbf{63.97}} & \textcolor[HTML]{056f28}{\textbf{91.92}} & \textcolor[HTML]{056f28}{\textbf{12.867}} \\
        \hdashline
        \multirow{3}{*}{UIUNet\cite{UIUNet}} & DLN Full & 81.64 & 84.56 & 97.21 & 13.838 & 60.24 & 60.13 & 87.83 & 54.675 & 90.61 & 91.30 & 99.56 & 7.724 & 69.66 & 63.29 & 90.24 & 14.955 \\
         & DLN Course + PAL & 37.24 & 39.02 & 71.50 & 32.099 & 36.93 & 39.19 & 84.41 & 108.115 & 75.5 & 76.35 & 99.05 & 6.894 & 52.98 & 51.60 & 92.26 & 33.08 \\
         & Ours & \textcolor[HTML]{056f28}{\textbf{74.01 (91\%)}} & \textcolor[HTML]{056f28}{\textbf{77.20}} & \textcolor[HTML]{056f28}{\textbf{96.68}} & \textcolor[HTML]{056f28}{\textbf{22.548}} & \textcolor{red}{\textbf{73.19 (121\%)}} & \textcolor[HTML]{056f28}{\textbf{71.17}} & \textcolor[HTML]{056f28}{\textbf{93.54}} & \textcolor[HTML]{056f28}{\textbf{18.316}} & \textcolor[HTML]{056f28}{\textbf{76.20 (84\%)}} & \textcolor[HTML]{056f28}{\textbf{77.29}} & \textcolor[HTML]{056f28}{\textbf{99.61}} & \textcolor[HTML]{056f28}{\textbf{6.752}} & \textcolor[HTML]{056f28}{\textbf{68.65 (99\%)}} & \textcolor[HTML]{056f28}{\textbf{62.18}} & \textcolor[HTML]{056f28}{\textbf{87.88}} & \textcolor[HTML]{056f28}{\textbf{13.759}} \\
        \hdashline
        \multirow{3}{*}{MSDA-Net\cite{MSDANet}} & DLN Full & 83.52 & 86.44 & 97.61 & 20.824 & 67.89 & 69.12 & 92.02 & 40.818 & 91.27 & 91.55 & 98.41 & 3.217 & 70.87 & 65.11 & 92.93 & 14.291 \\
         & DLN Course + PAL & 68.01 & 70.74 & 96.21 & 30.654 & 50.5 & 52.70 & 81.37 & 21.541 & 71.44 & 73.03 & 98.10 & 4.596 & 46.65 & 50.11 & 88.89 & 29.531 \\
         & Ours & \textcolor[HTML]{056f28}{\textbf{73.28 (88\%)}} & \textcolor[HTML]{056f28}{\textbf{76.28}} & \textcolor[HTML]{056f28}{\textbf{96.01}} & \textcolor[HTML]{056f28}{\textbf{25.004}} & \textcolor[HTML]{056f28}{\textbf{64.02 (94\%)}} & \textcolor[HTML]{056f28}{\textbf{63.99}} & \textcolor[HTML]{056f28}{\textbf{90.87}} & \textcolor[HTML]{056f28}{\textbf{30.596}} & \textcolor[HTML]{056f28}{\textbf{75.10 (82\%)}} & \textcolor[HTML]{056f28}{\textbf{76.37}} & \textcolor[HTML]{056f28}{\textbf{98.37}} & \textcolor[HTML]{056f28}{\textbf{5.622}} & \textcolor[HTML]{056f28}{\textbf{65.67 (93\%)}} & \textcolor[HTML]{056f28}{\textbf{64.58}} & \textcolor[HTML]{056f28}{\textbf{92.26}} & \textcolor[HTML]{056f28}{\textbf{23.42}} \\
        \hline
    \end{tabular}
    }
\end{table*}
\subsection{Stage Two: Knowledge Transfer for Lightweight Deployment}
While the expert teacher model from Stage One solves the Challenges 1 and 2, its massive size make it completely unsuitable for practical deployment (Challenge 3). Therefore, we designed Stage Two, with the target of transferring the knowledge from the upstream teacher model into a lightweight, efficient, and prompt-free downstream model $\mathcal{S}_{\psi}$.

\subsubsection{Pseudo Label Generation by Expert Teacher}
We employ the expert teacher to generate pseudo-masks $M_j^{pseudo}$ for the entire dataset $D_{all} = D_L \cup D_U$, treating even labeled samples as unlabeled by disregarding their ground truths. For each image $I_{j}$, we perform a forward pass to obtain the pseudo label:
\begin{equation}
  M_{j}^{pseudo}=\mathcal{E}_{\Phi}(I_{j};\mathcal{A}_{\theta^{*}}) 
\end{equation}
Thus, we construct a large pseudo label dataset $D_{pseudo}=(I_{j},M_{j}^{pseudo})_{j=1}^{M}$, which will serve as the sole source of supervision for the downstream model.

\subsubsection{Downstream Model Training}
After obtaining the pseudo label dataset $D_{pseudo}$, we select a lightweight network $\mathcal{S}_{\psi}$ suitable for IR-SOT as the downstream student model. We train this downstream model $\mathcal{S}_{\psi}$ from scratch, enabling it to learn to fit the pseudo labels generated by the expert teacher.

\subsubsection{Optimization Objective Update}
During the training process of the second stage, we only use the main task loss $\mathcal{L}_{main}$. This loss function is designed to minimize the difference between the predicted mask $M_{pred}$ of the downstream model $\mathcal{S}_{\psi}$ and the pseudo label $M_{pseudo}$ generated by the expert teacher. As in stage one, $\mathcal{L}_{main}$ is the weighted sum of BCE and Dice loss:
\begin{equation}
\resizebox{.85\hsize}{!}{$
\mathcal{L}_{main} = \lambda_{bce} \cdot \mathcal{L}_{BCE}(M_{pred}, M_{pseudo}) + \lambda_{dice} \cdot \mathcal{L}_{Dice}(M_{pred}, M_{pseudo})
$}
\end{equation}
where $M_{pseudo} \in D_{pseudo}$. It is worth noting that the auxiliary sparsity loss $\mathcal{L}_{sparse}$ is no longer used here
, as the MoE adapter does not participate in the training of this stage.

\section{Experiment}

\subsection{Implementation Details}
We conduct experiments on four public datasets: SIRST3 \cite{SIRST3}, NUAA-SIRST \cite{NUAA-SIRST}, NUDT-SIRST \cite{NUDT-SIRST}, and IRSTD-1K \cite{IRSTD-1K}. We employ standard metrics $nIoU$, $mIoU$, $P_d$, $F_a$ and our performance recovery rate : the performance ratio of our 10\% data method against the 100\% data baseline—to quantify data efficiency.

We compare our two-stage semi-supervised paradigm (using 10\% data) against two key baselines: (1) the fully supervised paradigm as a theoretical upper bound, and (2) the SOTA point-supervised paradigm, PAL\cite{PAL}.

We apply these paradigms to eight SOTA model (ACM \cite{NUAA-SIRST}, ALCNet \cite{ALCNet}, MLCL-Net \cite{MLCLNet}, ALCL-Net \cite{ALCLNet}, DNANet \cite{NUDT-SIRST}, GGL-Net \cite{GGLNet}, UIUNet \cite{UIUNet}, and MSDA-Net \cite{MSDANet}). We use PyTorch with a ViT-h SAM backbone and train on 8$\times$A100-80G GPUs. All implementation details and hyperparameters are provided in the Supplementary Material.
\begin{figure*}[htbp]
    \centering
    \includegraphics[width=0.95\textwidth]{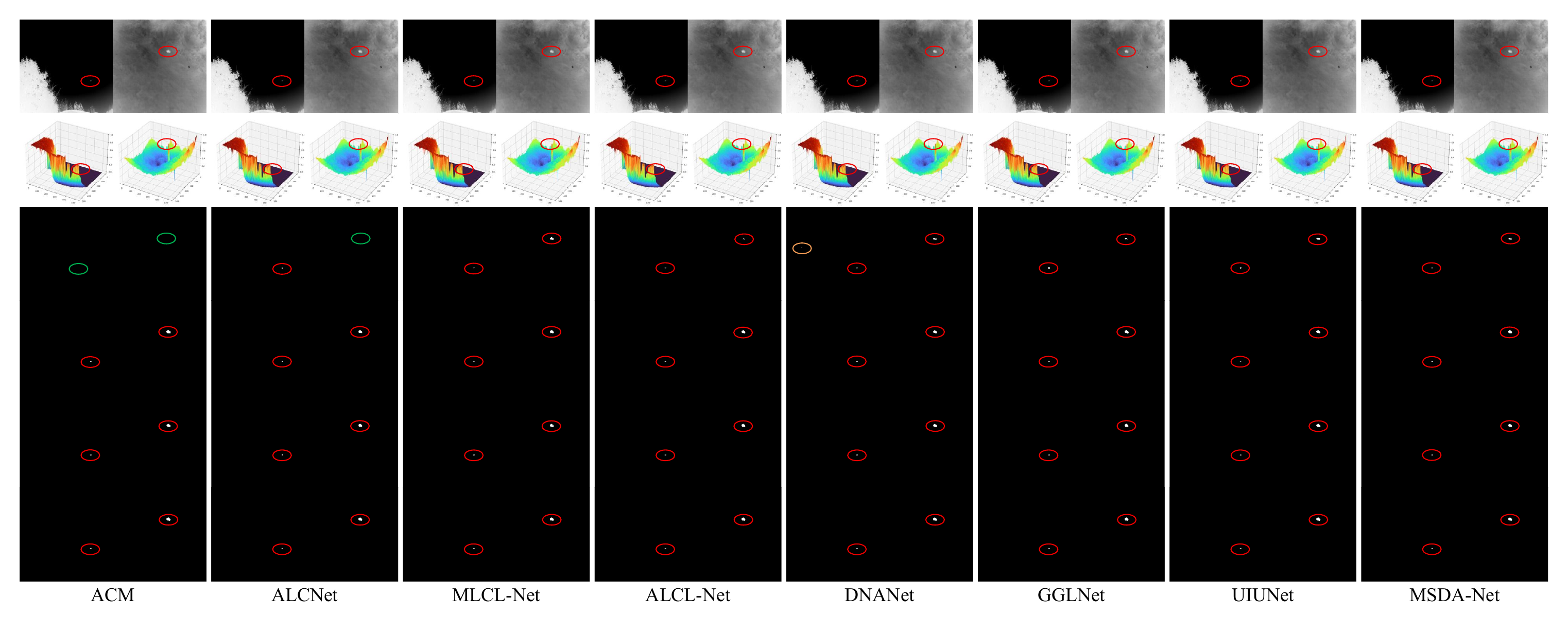}
    \caption{Qualitative comparison on the IRSTD-1K dataset. Each row from top to bottom displays: the original infrared image, the 3D brightness map of the original image, the results of PAL,  the results of the fully supervised method, the results of our method, and the ground truth (GT). For the predictions in this figure, \textcolor{red}{red} represents true positives (TP), \textcolor[HTML]{f59d55}{orange} represents false positives (FP), and \textcolor[HTML]{00ae4f}{green} represents false negatives (FN). The visual results show that our method significantly outperforms PAL in terms of performance and achieves results that are close to, or even surpass, those of the fully supervised method.}
    \label{fig:results}
\end{figure*}
\subsection{Results}
\subsubsection{Validation of Paradigm Necessity}

To validate our two-stage design, we benchmark against two pivotal baselines in Table \ref{tab:paradigm}: (A) Direct Training on 10\% Data: Training the student from scratch using only the 10\% labeled subset ($D_L$) leads to significant performance drops. This confirms that 10\% data is statistically insufficient for learning robust representations, inevitably causing overfitting. Our method circumvents this by using the expert teacher to propagate knowledge to unlabeled data, effectively amplifying the supervision signal. (B) Teacher Inference: Remarkably, our lightweight student matches or even surpasses the heavy Scalpel-SAM teacher. This demonstrates successful knowledge transfer: the student effectively internalizes the teacher's complex physical priors into a compact representation, resolving the dilemma between model capability and deployment efficiency.

\subsubsection{Comparison with Fully Supervised Methods}
\label{sec:exp_fully}
As quantified in Table \ref{tab:comparison}, our paradigm demonstrates exceptional data efficiency. Surprisingly, using only 10\% of labeled data, our method not only approaches but also surpasses the performance of fully supervised counterparts. \textbf{Notably, the performance under fully supervised training is often regarded as an upper bound. To date, no other semi-supervised paradigm has approached, let alone exceeded this fully supervised baseline}. We attribute this breakthrough to two factors: (1) Leveraging SAM's robust pre-trained knowledge instead of training from scratch on scarce data; and (2) The MoE Adapter's physical priors, which generate cleaner pseudo-labels by mitigating manual annotation noise. In addition, in Figure \ref{fig:results}, we present qualitative comparisons of segmentation results produced by different methods.
\begin{table*}[htbp]
    \centering
    \tiny
    \caption{Ablation study on the necessity of the two-stage paradigm. Our complete paradigm significantly outperforms two key baselines, demonstrating the immense value of first distilling an expert teacher and then performing transfer.}
    \label{tab:paradigm}
    \resizebox{\textwidth}{!}{%
    \begin{tabular}{llcccccccccccccccc}
        \hline
        \multirow{2}{*}{Scheme} & \multirow{2}{*}{Description} & \multicolumn{4}{c}{SIRST3-Test} & \multicolumn{4}{c}{NUAA-SIRST-Test} & \multicolumn{4}{c}{NUDT-SIRST-Test} & \multicolumn{4}{c}{IRSTD-1K-Test} \\
        \cline{3-18}
         & & $mIoU$ & $nIoU$ & $P_d$ & $F_a$ & $mIoU$ & $nIoU$ & $P_d$ & $F_a$ & $mIoU$ & $nIoU$ & $P_d$ & $F_a$ & $mIoU$ & $nIoU$ & $P_d$ & $F_a$ \\
        \hline
        ACM\cite{NUAA-SIRST} & DLN Semi & * & * & * & * & * & * & * & * & 25.61 & 23.15 & 51.64 & 83.188 & 51.96 & 48.61 & 90.24 & 68.475 \\
        ALCNet\cite{ALCNet} & DLN Semi & 2.17 & 1.2 & 5.05 & 3.042 & 64.30 & 63.09 & 88.59 & 29.224 & 24.80 & 23.72 & 72.38 & 94.586 & 9.26 & 5.68 & 23.57 & 32.036 \\
        MLCL-Net\cite{MLCLNet} & DLN Semi & 53.63 & 59.70 & 86.84 & 64.991 & 72.44 & 72.34 & 95.44 & 21.746 & 48.60 & 57.21 & 83.49 & 84.429 & 60.58 & 59.46 & 93.27 & 32.947 \\
        ALCL-Net\cite{ALCLNet} & DLN Semi & 48.80 & 51.61 & 83.32 & 51.966 & 65.27 & 65.87 & 90.49 & 36.564 & 44.28 & 51.39 & 82.01 & 118.554 & 57.98 & 53.80 & 90.57 & 67.184 \\
        DNANet\cite{NUDT-SIRST} & DLN Semi & 52.72 & 59.57 & 85.12 & 61.670 & 71.04 & 72.44 & 95.06 & 50.559 & 44.18 & 51.69 & 80.63 & 106.72 & 59.42 & 57.96 & 87.21 & 30.935 \\
        GGL-Net\cite{GGLNet} & DLN Semi & 41.03 & 45.12 & 83.06 & 104.618 & 69.78 & 70.43 & 95.06 & 34.781 & 48.57 & 56.57 & 84.76 & 131.699 & 56.44 & 55.71 & 91.58 & 42.816 \\
        UIUNet\cite{UIUNet} & DLN Semi & 56.23 & 60.38 & 86.64 & 70.750 & 68.54 & 65.99 & 93.16 & 26.960 & 46.73 & 53.10 & 80.95 & 84.59 & 59.69 & 56.38 & 94.61 & 64.774 \\
        MSDA-Net\cite{MSDANet} & DLN Semi & 48.60 & 53.27 & 83.06 & 68.512 & 61.62 & 63.02 & 87.45 & 39.377 & 44.65 & 53.66 & 80.21 & 106.582 & 62.81 & 59.32 & 92.26 & 28.658 \\
        Ours (B) & DLN Semi & 68.02 & 78.51 & 94.53 & 95.500 & 64.05 & 74.58 & 91.02 & 91.05 & 70.10 & 81.49 & 95.82 & 96.80 & 65.67 & 75.21 & 92.26 & 90.10 \\
        Ours(UIUNet)& DLN Semi & \textcolor[HTML]{056f28}{\textbf{74.01}} & \textcolor[HTML]{056f28}{\textbf{77.20}} & \textcolor[HTML]{056f28}{\textbf{96.68}} & \textcolor[HTML]{056f28}{\textbf{22.548}} & \textcolor[HTML]{056f28}{\textbf{73.19}} & \textcolor[HTML]{056f28}{\textbf{71.17}} & \textcolor[HTML]{056f28}{\textbf{93.54}} & \textcolor[HTML]{056f28}{\textbf{18.316}} & \textcolor[HTML]{056f28}{\textbf{72.20}} & \textcolor[HTML]{056f28}{\textbf{74.29}} & \textcolor[HTML]{056f28}{\textbf{96.61}} & \textcolor[HTML]{056f28}{\textbf{16.752}} & \textcolor[HTML]{056f28}{\textbf{68.65}} & \textcolor[HTML]{056f28}{\textbf{62.18}} & \textcolor[HTML]{056f28}{\textbf{87.88}} & \textcolor[HTML]{056f28}{\textbf{13.759}} \\
        \hline
    \end{tabular}%
    }  
\end{table*}

\label{sec:exp_ablation}
\begin{table*}[htbp]
\centering
\tiny
\caption{A comprehensive ablation study on the routing mechanism, expert design, and adapter insertion strategy.}
\label{tab:comprehensive_ablation}
\resizebox{\textwidth}{!}{%
\begin{tabular}{llcccccccccccccccc}
\hline
\multirow{2}{*}{} & \multirow{2}{*}{Variant} & \multicolumn{4}{c}{SIRST3-Test} & \multicolumn{4}{c}{NUAA-SIRST-Test} & \multicolumn{4}{c}{NUDT-SIRST-Test} & \multicolumn{4}{c}{IRSTD-1K-Test} \\
\cline{3-18}
& & $mIoU$ & $nIoU$ & $P_d$ & $F_a$ & $mIoU$ & $nIoU$ & $P_d$ & $F_a$ & $mIoU$ & $nIoU$ & $P_d$ & $F_a$ & $mIoU$ & $nIoU$ & $P_d$ & $F_a$ \\
\hline
\multirow{2}{*}{MoE Routing} & Static (Avg. Fusion) & 64.58 & 75.05 & 91.91 & 97.95 & 61.23 & 71.88 & 88.55 & 92.98 & 66.92 & 78.48 & 93.25 & 98.01 & 62.81 & 72.61 & 89.85 & 92.11 \\
& \textbf{Dynamic (Ours)} & \textcolor[HTML]{056f28}{\textbf{68.02}} & \textcolor[HTML]{056f28}{\textbf{78.51}} & \textcolor[HTML]{056f28}{\textbf{95.50}} & \textcolor[HTML]{056f28}{\textbf{94.53}} & \textcolor[HTML]{056f28}{\textbf{64.05}} & \textcolor[HTML]{056f28}{\textbf{74.58}} & \textcolor[HTML]{056f28}{\textbf{91.05}} & \textcolor[HTML]{056f28}{\textbf{91.02}} & \textcolor[HTML]{056f28}{\textbf{70.10}} & \textcolor[HTML]{056f28}{\textbf{81.49}} & \textcolor[HTML]{056f28}{\textbf{96.80}} & \textcolor[HTML]{056f28}{\textbf{95.82}} & \textcolor[HTML]{056f28}{\textbf{65.67}} & \textcolor[HTML]{056f28}{\textbf{75.21}} & \textcolor[HTML]{056f28}{\textbf{90.10}} & \textcolor[HTML]{056f28}{\textbf{92.26}} \\
\hdashline
\multirow{2}{*}{Expert Design} 
& SAM-Adapter & 62.13 & 72.88 & 90.05 & 98.51 & 59.98 & 70.35 & 87.21 & 94.13 & 65.25 & 76.91 & 92.07 & 98.85 & 61.39 & 71.05 & 88.52 & 93.04 \\
& LoRA(r = 8) & 65.52 & 76.54 & 93.61 & 95.92 & 62.48 & 73.01 & 89.54 & 92.15 & 68.31 & 79.88 & 94.63 & 97.18 & 64.03 & 73.77 & 90.88 & 91.04 \\
& Black-box & 63.45 & 74.01 & 91.52 & 97.33 & 60.71 & 71.24 & 87.95 & 93.58 & 66.19 & 77.80 & 92.74 & 98.42 & 62.18 & 71.89 & 89.21 & 92.65 \\
& \textbf{White-box (Ours)} & \textcolor[HTML]{056f28}{\textbf{68.02}} & \textcolor[HTML]{056f28}{\textbf{78.51}} & \textcolor[HTML]{056f28}{\textbf{95.50}} & \textcolor[HTML]{056f28}{\textbf{94.53}} & \textcolor[HTML]{056f28}{\textbf{64.05}} & \textcolor[HTML]{056f28}{\textbf{74.58}} & \textcolor[HTML]{056f28}{\textbf{91.05}} & \textcolor[HTML]{056f28}{\textbf{91.02}} & \textcolor[HTML]{056f28}{\textbf{70.10}} & \textcolor[HTML]{056f28}{\textbf{81.49}} & \textcolor[HTML]{056f28}{\textbf{96.80}} & \textcolor[HTML]{056f28}{\textbf{95.82}} & \textcolor[HTML]{056f28}{\textbf{65.67}} & \textcolor[HTML]{056f28}{\textbf{75.21}} & \textcolor[HTML]{056f28}{\textbf{90.10}} & \textcolor[HTML]{056f28}{\textbf{92.26}} \\
\hline
\end{tabular}%
}
\end{table*}
\subsubsection{Ablation Study}
\paragraph{Advancement of MoE routing and white-box expert Design}
Table \ref{tab:comprehensive_ablation} validates our core adapter design. The results confirm two key points: (1) Dynamic routing significantly outperforms the static average fusion baseline; and (2) our white-box experts substantially outperform generic black-box PEFT baselines (SAM-Adapter, LoRA). This proves the fundamental advantage of our domain-driven design.
\begin{table*}[h!]
\tiny
\centering
\caption{Analysis of the white-box expert combination effectiveness.}
\label{tab:expert_combination}
\resizebox{\textwidth}{!}{%
\begin{tabular}{lcccccccccccccccc}
\hline
\multirow{2}{*}{Variant} & \multicolumn{4}{c}{SIRST3-Test} & \multicolumn{4}{c}{NUAA-SIRST-Test} & \multicolumn{4}{c}{NUDT-SIRST-Test} & \multicolumn{4}{c}{IRSTD-1K-Test} \\
\cline{2-17}
& $mIoU$ & $nIoU$ & $P_d$ & $F_a$ & $mIoU$ & $nIoU$ & $P_d$ & $F_a$ & $mIoU$ & $nIoU$ & $P_d$ & $F_a$ & $mIoU$ & $nIoU$ & $P_d$ & $F_a$ \\
\hline
PI & 62.15 & 72.93 & 98.55 & 90.18 & 58.39 & 68.95 & 95.31 & 85.78 & 63.92 & 75.15 & 98.99 & 91.53 & 60.23 & 69.91 & 94.19 & 87.25 \\
SPD & 62.51 & 73.28 & 98.21 & 90.45 & 58.77 & 69.31 & 94.98 & 86.03 & 64.29 & 75.48 & 98.73 & 91.88 & 60.65 & 70.32 & 93.88 & 87.61 \\
HP & 61.98 & 72.77 & 98.73 & 90.01 & 58.11 & 68.69 & 95.67 & 85.51 & 63.65 & 74.88 & 99.15 & 91.29 & 59.99 & 69.65 & 94.43 & 86.99 \\
TG & 63.02 & 73.75 & 97.88 & 90.91 & 59.13 & 69.72 & 94.51 & 86.49 & 64.81 & 76.01 & 98.41 & 92.25 & 61.12 & 70.79 & 93.45 & 88.04 \\
\hdashline
PI + SPD & 65.38 & 75.85 & 96.93 & 92.61 & 61.31 & 71.98 & 92.95 & 88.68 & 67.05 & 78.67 & 97.88 & 93.84 & 63.07 & 72.85 & 91.99 & 90.01 \\
PI + HP & 65.11 & 75.59 & 97.15 & 92.40 & 61.02 & 71.69 & 93.24 & 88.41 & 66.79 & 78.38 & 98.13 & 93.59 & 62.81 & 72.58 & 92.21 & 89.77 \\
PI + TG & 65.93 & 76.42 & 96.58 & 92.99 & 61.88 & 72.51 & 92.46 & 89.15 & 67.65 & 79.28 & 97.49 & 94.31 & 63.63 & 73.39 & 91.43 & 90.52 \\
SPD + HP & 65.45 & 75.93 & 96.89 & 92.68 & 61.39 & 72.05 & 92.88 & 88.75 & 67.13 & 78.74 & 97.81 & 93.91 & 63.15 & 72.93 & 91.91 & 90.08 \\
SPD + TG & 66.25 & 76.78 & 96.31 & 93.28 & 62.19 & 72.81 & 92.11 & 89.43 & 67.99 & 79.59 & 97.23 & 94.59 & 63.94 & 73.68 & 91.12 & 90.81 \\
HP + TG & 65.80 & 76.31 & 96.69 & 92.89 & 61.73 & 72.35 & 92.59 & 89.01 & 67.51 & 79.13 & 97.60 & 94.19 & 63.51 & 73.27 & 91.58 & 90.41 \\
\hdashline
PI + SPD + HP & 67.21 & 77.72 & 95.88 & 94.09 & 63.29 & 73.81 & 91.49 & 90.48 & 69.11 & 80.64 & 97.01 & 95.35 & 64.99 & 74.59 & 90.51 & 91.81 \\
PI + HP + TG & 67.59 & 78.08 & 95.65 & 94.31 & 63.65 & 74.19 & 91.18 & 90.79 & 69.53 & 81.01 & 96.85 & 95.59 & 65.31 & 74.89 & 90.29 & 92.05 \\
\hdashline
All & \textcolor[HTML]{056f28}{\textbf{68.02}} & \textcolor[HTML]{056f28}{\textbf{78.51}} & \textcolor[HTML]{056f28}{\textbf{95.50}} & \textcolor[HTML]{056f28}{\textbf{94.53}} & \textcolor[HTML]{056f28}{\textbf{64.05}} & \textcolor[HTML]{056f28}{\textbf{74.58}} & \textcolor[HTML]{056f28}{\textbf{91.05}} & \textcolor[HTML]{056f28}{\textbf{91.02}} & \textcolor[HTML]{056f28}{\textbf{70.10}} & \textcolor[HTML]{056f28}{\textbf{81.49}} & \textcolor[HTML]{056f28}{\textbf{96.80}} & \textcolor[HTML]{056f28}{\textbf{95.82}} & \textcolor[HTML]{056f28}{\textbf{65.67}} & \textcolor[HTML]{056f28}{\textbf{75.21}} & \textcolor[HTML]{056f28}{\textbf{90.10}} & \textcolor[HTML]{056f28}{\textbf{92.26}} \\
\hline
\end{tabular}%
}
\end{table*}

\begin{table*}[htbp]
\centering
\tiny
\caption{Ablation study on the Adapter insertion strategy.}
\label{tab:supp_insertion}
\resizebox{\textwidth}{!}{%
\begin{tabular}{llcccccccccccccccc}
\hline
\multirow{2}{*}{} & \multirow{2}{*}{Variant} & \multicolumn{4}{c}{SIRST3-Test} & \multicolumn{4}{c}{NUAA-SIRST-Test} & \multicolumn{4}{c}{NUDT-SIRST-Test} & \multicolumn{4}{c}{IRSTD-1K-Test} \\
\cline{3-18}
& & $mIoU$ & $nIoU$ & $P_d$ & $F_a$ & $mIoU$ & $nIoU$ & $P_d$ & $F_a$ & $mIoU$ & $nIoU$ & $P_d$ & $F_a$ & $mIoU$ & $nIoU$ & $P_d$ & $F_a$ \\
\hline
\multirow{6}{*}{Adapter Insertion}
& Last 1/2 ViT Layers & 66.98 & 77.51 & 93.95 & 95.99 & 62.91 & 73.37 & 90.08 & 91.88 & 68.85 & 80.29 & 94.91 & 97.34 & 64.48 & 74.09 & 91.35 & 90.95 \\
& First 1/2 ViT Layers & 65.55 & 76.16 & 92.77 & 96.81 & 61.85 & 72.28 & 88.98 & 92.71 & 67.57 & 79.05 & 93.68 & 98.01 & 63.53 & 73.21 & 90.37 & 91.78 \\
& All ViT Layers & 60.57 & 71.15 & 88.62 & 99.53 & 58.72 & 68.85 & 85.62 & 95.93 & 63.96 & 75.32 & 90.47 & 99.34 & 60.01 & 69.55 & 86.87 & 94.21 \\
& Last 4 ViT Layers & 67.67 & 78.15 & 94.28 & 95.71 & 63.62 & 74.11 & 90.67 & 91.35 & 69.69 & 81.06 & 95.45 & 96.99 & 65.25 & 74.87 & 91.97 & 90.43 \\
& None & 35.12 & 48.29 & 75.33 & 132.15 & 33.48 & 45.91 & 71.88 & 158.29 & 38.21 & 50.78 & 78.41 & 141.33 & 34.03 & 46.59 & 73.51 & 162.51 \\
& \textbf{Last 2 ViT Layers (Ours)} & \textcolor[HTML]{056f28}{\textbf{68.02}} & \textcolor[HTML]{056f28}{\textbf{78.51}} & \textcolor[HTML]{056f28}{\textbf{95.50}} & \textcolor[HTML]{056f28}{\textbf{94.53}} & \textcolor[HTML]{056f28}{\textbf{64.05}} & \textcolor[HTML]{056f28}{\textbf{74.58}} & \textcolor[HTML]{056f28}{\textbf{91.05}} & \textcolor[HTML]{056f28}{\textbf{91.02}} & \textcolor[HTML]{056f28}{\textbf{70.10}} & \textcolor[HTML]{056f28}{\textbf{81.49}} & \textcolor[HTML]{056f28}{\textbf{96.80}} & \textcolor[HTML]{056f28}{\textbf{95.82}} & \textcolor[HTML]{056f28}{\textbf{65.67}} & \textcolor[HTML]{056f28}{\textbf{75.21}} & \textcolor[HTML]{056f28}{\textbf{90.10}} & \textcolor[HTML]{056f28}{\textbf{92.26}} \\
\hline
\end{tabular}%
}
\end{table*}

\begin{table*}[htbp]
\tiny
\centering
\caption{Ablation experiment on the sparsity weight $\lambda_{sparse}$.}
\label{tab:supp_sparse_weight}
\resizebox{\textwidth}{!}{
\begin{tabular}{lcccccccccccccccc}
\hline
\multirow{2}{*}{$\lambda_{sparse}$} & \multicolumn{4}{c}{SIRST3-Test} & \multicolumn{4}{c}{NUAA-SIRST-Test} & \multicolumn{4}{c}{NUDT-SIRST-Test} & \multicolumn{4}{c}{IRSTD-1K-Test} \\
\cline{2-17}
& $mIoU$ & $nIoU$ & $P_d$ & $F_a$ & $mIoU$ & $nIoU$ & $P_d$ & $F_a$ & $mIoU$ & $nIoU$ & $P_d$ & $F_a$ & $mIoU$ & $nIoU$ & $P_d$ & $F_a$ \\
\hline
0.0010 & 67.18 & 77.65 & 96.45 & 93.81 & 63.13 & 73.69 & 93.61 & 90.15 & 69.19 & 80.59 & 96.98 & 94.98 & 64.73 & 74.25 & 91.89 & 91.31 \\
0.0012 & 67.35 & 77.83 & 96.38 & 93.99 & 63.31 & 73.85 & 93.52 & 90.35 & 69.38 & 80.79 & 96.90 & 95.19 & 64.92 & 74.48 & 91.80 & 91.55 \\
0.0014 & 67.51 & 78.00 & 96.30 & 94.13 & 63.50 & 74.05 & 93.43 & 90.56 & 69.59 & 81.01 & 96.81 & 95.40 & 65.13 & 74.70 & 91.71 & 91.78 \\
0.0016 & 67.69 & 78.18 & 96.22 & 94.28 & 63.71 & 74.24 & 93.34 & 90.75 & 69.78 & 81.20 & 96.73 & 95.59 & 65.34 & 74.91 & 91.62 & 92.00 \\
0.0018 & 67.85 & 78.35 & 96.13 & 94.41 & 63.89 & 74.42 & 93.24 & 90.89 & 69.95 & 81.37 & 96.67 & 95.72 & 65.52 & 75.08 & 91.51 & 92.15 \\
0.0020 & 67.93 & 78.43 & 96.08 & 94.47 & 63.97 & 74.50 & 93.19 & 90.95 & 70.03 & 81.43 & 96.64 & 95.77 & 65.60 & 75.15 & 91.46 & 92.21 \\
\hdashline
0.0040 & 67.98 & 78.47 & 96.06 & 94.50 & 64.01 & 74.54 & 93.17 & 90.98 & 70.06 & 81.46 & 96.63 & 95.79 & 65.63 & 75.18 & 91.44 & 92.23 \\
0.0060 & 68.01 & 78.50 & 96.05 & 94.52 & 64.03 & 74.56 & 93.16 & 91.00 & 70.08 & 81.48 & 96.62 & 95.81 & 65.65 & 75.20 & 91.43 & 92.25 \\
0.0080 & 68.02 & 78.51 & 96.04 & 94.53 & 64.04 & 74.57 & 93.15 & 91.01 & 70.09 & 81.49 & 96.62 & 95.82 & 65.66 & 75.21 & 91.42 & 92.26 \\
0.0100 (\textbf{Ours}) & \textcolor[HTML]{056f28}{\textbf{68.02}} & \textcolor[HTML]{056f28}{\textbf{78.51}} & \textcolor[HTML]{056f28}{\textbf{95.50}} & \textcolor[HTML]{056f28}{\textbf{94.49}} & \textcolor[HTML]{056f28}{\textbf{64.05}} & \textcolor[HTML]{056f28}{\textbf{74.58}} & \textcolor[HTML]{056f28}{\textbf{91.05}} & \textcolor[HTML]{056f28}{\textbf{91.02}} & \textcolor[HTML]{056f28}{\textbf{70.10}} & \textcolor[HTML]{056f28}{\textbf{81.49}} & \textcolor[HTML]{056f28}{\textbf{96.80}} & \textcolor[HTML]{056f28}{\textbf{95.81}} & \textcolor[HTML]{056f28}{\textbf{65.67}} & \textcolor[HTML]{056f28}{\textbf{75.21}} & \textcolor[HTML]{056f28}{\textbf{90.10}} & \textcolor[HTML]{056f28}{\textbf{92.24}} \\
\hdashline
0.0400 & 67.81 & 78.29 & 96.15 & 94.35 & 63.82 & 74.33 & 93.28 & 90.80 & 69.88 & 81.27 & 96.75 & 95.63 & 65.45 & 75.00 & 91.58 & 92.06 \\
0.0800 & 67.43 & 77.91 & 96.34 & 94.05 & 63.41 & 73.95 & 93.48 & 90.43 & 69.49 & 80.89 & 96.86 & 95.28 & 65.05 & 74.61 & 91.75 & 91.69 \\
\hline
\end{tabular}%
}
\end{table*}

\paragraph{Expert combination effectiveness analysis}
We ablated the combination of our four white-box experts, as shown in Table \ref{tab:expert_combination}). The results confirm their complementarity: removing any single expert degrades performance, while the full four-expert model (All) achieves the best results.

\paragraph{Ablation of adapter insertion layers}
\label{sec:supp_insertion}
We ablated the adapter insertion position and number, as shown in Table \ref{tab:supp_insertion}. The experiments confirm our design choice: inserting the MoE Adapter only after the last two ViT blocks yields the best balance of performance and cost. Notably, inserting adapters after all ViT layers performed the poorest, likely because disrupting SAM's foundational features at shallow layers.

\paragraph{Ablation of Sparsity Loss Weight (\texorpdfstring{$\lambda_{sparse}$}{Lsparse})}
\label{sec:supp_sparse}
We conducted an ablation experiment on the weight coefficient $\lambda_{sparse}$ of the auxiliary sparsity loss. As shown in Table \ref{tab:supp_sparse_weight}, the chosen value of $\lambda_{sparse} = 0.0100$ achieves the optimal performance balance across all datasets.

\section{Related Work}

Our work is closely related to IR-SOT \cite{zhao2022single} and PEFT \cite{kirkpatrick2017overcoming}. The mainstream methods of IR-SOT have evolved from early local contrast measures (LCM) \cite{MPCM} to deep learning-based segmentation networks (such as DNANet \cite{NUDT-SIRST} and UIUNet \cite{wu2022uiu}). However, most of these methods follow the training from scratch paradigm, failing to leverage the priors of large foundational models and heavily relying on expensive pixel-level annotations. On the other hand, PEFT techniques have been successfully applied to adapt SAM \cite{chen2023sam}. However, these methods are generally universal black boxes that do not account for the unique physical properties of infrared images during adaptation.

\section{Conclusion}
We proposed a two-stage knowledge distillation and transfer paradigm to adapt foundation models to specialized, data-scarce domains. This paradigm first refines SAM into an expert teacher (Scalpel-SAM) using our novel Hierarchical MoE Adapter, which is composed of four white-box physical operators. It then transfers this knowledge to a lightweight, prompt-free downstream model. Experiments show this achieves or surpasses fully supervised SOTA performance with only 10\% of labeled data. More importantly, the domain-principle-driven PEFT methodology we propose offers a promising pathway for extending the capabilities of foundational models to other specialized domains, such as medical imaging.

{
    \tiny
    \bibliographystyle{IEEEbib}
    \bibliography{icme2026references}
}

\clearpage
\setcounter{page}{1}
\setcounter{section}{0}
\renewcommand{\thesection}{\Alph{section}}
\begin{center}
{\Large\bfseries Supplementary Material for}\\[0.6em]
{\large\bfseries Scalpel-SAM: Precision Adaptation of SAM via }\\[0.5em]
{\large\bfseries Domain-Principle-Driven Expert Adapters}
\end{center}
\vspace{2em}

\section{Supplementary Details for Methodology}
\label{sec:supp_method}
This document provides detailed formulas and implementation details omitted for brevity in the main paper \S 3 (Methodology).

\subsection{Hierarchical MoE Adapter}
\label{sec:supp_adapter}
\paragraph{Adapter Integration}
For the $l$-th layer of the ViT encoder, the input features are $x_l$, and the output $x_{l+1}$ after passing through the Transformer module $\mathcal{B}l$ and our adapter $\mathcal{A}^{(l)}$ is:
\begin{equation}
\label{eq:supp_adapter_out}
x_{l+1} = \mathcal{B}_l(x_l) + \mathcal{A}^{(l)}(x_l)
\end{equation}
\begin{table*}[htbp]
\tiny
\centering
\caption{Analysis of the Effectiveness of the 'White-Box' Expert Combination. The full four-expert combination yields the best performance, and removing any individual expert leads to a performance decline, demonstrating the complementarity of our design.}
\label{tab:expert_combination}
\resizebox{\textwidth}{!}{
\begin{tabular}{lcccccccccccccccc}
\hline
\multirow{2}{*}{Variant} & \multicolumn{4}{c}{SIRST3-Test} & \multicolumn{4}{c}{NUAA-SIRST-Test} & \multicolumn{4}{c}{NUDT-SIRST-Test} & \multicolumn{4}{c}{IRSTD-1K-Test} \\
\cline{2-17}
& $IoU$ & $nIoU$ & $P_d$ & $F_a$ & $IoU$ & $nIoU$ & $P_d$ & $F_a$ & $IoU$ & $nIoU$ & $P_d$ & $F_a$ & $IoU$ & $nIoU$ & $P_d$ & $F_a$ \\
\hline
PI & 62.15 & 72.93 & 98.55 & 90.18 & 58.39 & 68.95 & 95.31 & 85.78 & 63.92 & 75.15 & 98.99 & 91.53 & 60.23 & 69.91 & 94.19 & 87.25 \\
SPD & 62.51 & 73.28 & 98.21 & 90.45 & 58.77 & 69.31 & 94.98 & 86.03 & 64.29 & 75.48 & 98.73 & 91.88 & 60.65 & 70.32 & 93.88 & 87.61 \\
HP & 61.98 & 72.77 & 98.73 & 90.01 & 58.11 & 68.69 & 95.67 & 85.51 & 63.65 & 74.88 & 99.15 & 91.29 & 59.99 & 69.65 & 94.43 & 86.99 \\
TG & 63.02 & 73.75 & 97.88 & 90.91 & 59.13 & 69.72 & 94.51 & 86.49 & 64.81 & 76.01 & 98.41 & 92.25 & 61.12 & 70.79 & 93.45 & 88.04 \\
\hdashline
PI + SPD & 65.38 & 75.85 & 96.93 & 92.61 & 61.31 & 71.98 & 92.95 & 88.68 & 67.05 & 78.67 & 97.88 & 93.84 & 63.07 & 72.85 & 91.99 & 90.01 \\
PI + HP & 65.11 & 75.59 & 97.15 & 92.40 & 61.02 & 71.69 & 93.24 & 88.41 & 66.79 & 78.38 & 98.13 & 93.59 & 62.81 & 72.58 & 92.21 & 89.77 \\
PI + TG & 65.93 & 76.42 & 96.58 & 92.99 & 61.88 & 72.51 & 92.46 & 89.15 & 67.65 & 79.28 & 97.49 & 94.31 & 63.63 & 73.39 & 91.43 & 90.52 \\
SPD + HP & 65.45 & 75.93 & 96.89 & 92.68 & 61.39 & 72.05 & 92.88 & 88.75 & 67.13 & 78.74 & 97.81 & 93.91 & 63.15 & 72.93 & 91.91 & 90.08 \\
SPD + TG & 66.25 & 76.78 & 96.31 & 93.28 & 62.19 & 72.81 & 92.11 & 89.43 & 67.99 & 79.59 & 97.23 & 94.59 & 63.94 & 73.68 & 91.12 & 90.81 \\
HP + TG & 65.80 & 76.31 & 96.69 & 92.89 & 61.73 & 72.35 & 92.59 & 89.01 & 67.51 & 79.13 & 97.60 & 94.19 & 63.51 & 73.27 & 91.58 & 90.41 \\
\hdashline
PI + SPD + HP & 67.21 & 77.72 & 95.88 & 94.09 & 63.29 & 73.81 & 91.49 & 90.48 & 69.11 & 80.64 & 97.01 & 95.35 & 64.99 & 74.59 & 90.51 & 91.81 \\
PI + HP + TG & 67.59 & 78.08 & 95.65 & 94.31 & 63.65 & 74.19 & 91.18 & 90.79 & 69.53 & 81.01 & 96.85 & 95.59 & 65.31 & 74.89 & 90.29 & 92.05 \\
\hdashline
All & \textcolor[HTML]{056f28}{\textbf{68.02}} & \textcolor[HTML]{056f28}{\textbf{78.51}} & \textcolor[HTML]{056f28}{\textbf{95.50}} & \textcolor[HTML]{056f28}{\textbf{94.53}} & \textcolor[HTML]{056f28}{\textbf{64.05}} & \textcolor[HTML]{056f28}{\textbf{74.58}} & \textcolor[HTML]{056f28}{\textbf{91.05}} & \textcolor[HTML]{056f28}{\textbf{91.02}} & \textcolor[HTML]{056f28}{\textbf{70.10}} & \textcolor[HTML]{056f28}{\textbf{81.49}} & \textcolor[HTML]{056f28}{\textbf{96.80}} & \textcolor[HTML]{056f28}{\textbf{95.82}} & \textcolor[HTML]{056f28}{\textbf{65.67}} & \textcolor[HTML]{056f28}{\textbf{75.21}} & \textcolor[HTML]{056f28}{\textbf{90.10}} & \textcolor[HTML]{056f28}{\textbf{92.26}} \\
\hline
\end{tabular}%
}
\end{table*}
\subsubsection{Dynamic Expert Routing Mechanism}
The routing network $\mathcal{R}(\cdot)$ receives the input features $x_l \in \mathbb{R}^{N \times C}$ of the layer and computes the gating weights $\mathbf{w} \in \mathbb{R}^K$ for the $K$ experts via a small MLP:
\begin{equation}
\label{eq:supp_routing_weights}
\mathbf{w} = \text{Softmax}(\mathbf{W}_2 \cdot \text{ReLU}(\mathbf{W}_1 \cdot \text{GAP}(x_l) + \mathbf{b}_1) + \mathbf{b}_2)
\end{equation}
where $\text{GAP}(\cdot)$ is the global average pooling operation. We employ a Soft Selection mechanism, where the outputs of all experts $\mathcal{E}i(x_l)$ are weighted and fused to obtain the final feature adjustment:
\begin{equation}
\label{eq:supp_expert_fusion}
    \mathcal{A}^{(l)}(x_l) = \sum_{i=1}^{K} w_i \cdot \mathcal{E}_i(x_l)
\end{equation}
\subsubsection{“White-Box” Expert Toolbox: Detailed Implementation}
\begin{table*}[htbp]
\centering
\tiny
\caption{Ablation study on the Adapter insertion strategy.}
\label{tab:supp_insertion}
\resizebox{\textwidth}{!}{%
\begin{tabular}{llcccccccccccccccc}
\hline
\multirow{2}{*}{} & \multirow{2}{*}{Variant} & \multicolumn{4}{c}{SIRST3-Test} & \multicolumn{4}{c}{NUAA-SIRST-Test} & \multicolumn{4}{c}{NUDT-SIRST-Test} & \multicolumn{4}{c}{IRSTD-1K-Test} \\
\cline{3-18}
& & mIoU & nIoU & $P_d$ & $F_a$ & mIoU & nIoU & $P_d$ & $F_a$ & mIoU & nIoU & $P_d$ & $F_a$ & mIoU & nIoU & $P_d$ & $F_a$ \\
\hline
\multirow{6}{*}{Adapter Insertion}
& Last 1/2 ViT Layers & 66.98 & 77.51 & 93.95 & 95.99 & 62.91 & 73.37 & 90.08 & 91.88 & 68.85 & 80.29 & 94.91 & 97.34 & 64.48 & 74.09 & 91.35 & 90.95 \\
& First 1/2 ViT Layers & 65.55 & 76.16 & 92.77 & 96.81 & 61.85 & 72.28 & 88.98 & 92.71 & 67.57 & 79.05 & 93.68 & 98.01 & 63.53 & 73.21 & 90.37 & 91.78 \\
& All ViT Layers & 60.57 & 71.15 & 88.62 & 99.53 & 58.72 & 68.85 & 85.62 & 95.93 & 63.96 & 75.32 & 90.47 & 99.34 & 60.01 & 69.55 & 86.87 & 94.21 \\
& Last 4 ViT Layers & 67.67 & 78.15 & 94.28 & 95.71 & 63.62 & 74.11 & 90.67 & 91.35 & 69.69 & 81.06 & 95.45 & 96.99 & 65.25 & 74.87 & 91.97 & 90.43 \\
& None & 35.12 & 48.29 & 75.33 & 132.15 & 33.48 & 45.91 & 71.88 & 158.29 & 38.21 & 50.78 & 78.41 & 141.33 & 34.03 & 46.59 & 73.51 & 162.51 \\
& \textbf{Last 2 ViT Layers (Ours)} & \textcolor[HTML]{056f28}{\textbf{68.02}} & \textcolor[HTML]{056f28}{\textbf{78.51}} & \textcolor[HTML]{056f28}{\textbf{95.50}} & \textcolor[HTML]{056f28}{\textbf{94.53}} & \textcolor[HTML]{056f28}{\textbf{64.05}} & \textcolor[HTML]{056f28}{\textbf{74.58}} & \textcolor[HTML]{056f28}{\textbf{91.05}} & \textcolor[HTML]{056f28}{\textbf{91.02}} & \textcolor[HTML]{056f28}{\textbf{70.10}} & \textcolor[HTML]{056f28}{\textbf{81.49}} & \textcolor[HTML]{056f28}{\textbf{96.80}} & \textcolor[HTML]{056f28}{\textbf{95.82}} & \textcolor[HTML]{056f28}{\textbf{65.67}} & \textcolor[HTML]{056f28}{\textbf{75.21}} & \textcolor[HTML]{056f28}{\textbf{90.10}} & \textcolor[HTML]{056f28}{\textbf{92.26}} \\
\hline
\end{tabular}%
}
\end{table*}
\paragraph{1. PIMDO}
The continuous form of the core idea of PIMDO is shown in Eq.~\eqref{eq:pimdo} in the main paper. In the implementation, we numerically discretize it using the explicit forward Euler method. For any pixel $x_{i,j}$, its update in the $t+1$ iteration is given by:
\begin{equation}
\label{eq:supp_pimdo}
x^{t+1}_{i,j} = x^{t}_{i,j} + \Delta t \cdot \nabla \cdot (c(|\nabla x^t_{i,j}|) \nabla x^t_{i,j})
\end{equation}
where the time step $\Delta t$ is set to 0.2, and the gradient $\nabla x$ is approximated using a fixed 3x3 Sobel operator. This evolution process is iterated $T=3$ times.

\paragraph{2. SPD}
Eq.~\eqref{eq:spd} in the main paper shows the reconstruction process of SPD. The analysis filter bank $\mathcal{W}_{ana}$ consists of a set of parallel, learnable 2D depthwise separable convolutions, which decompose the input features into $J=4$ different frequency subspaces.

\paragraph{3. HPLSM}
Eq.~\eqref{eq:hplsm} in the main paper shows the modulation process of HPLSM. The hypernetwork $\mathcal{H}_{\psi}(\cdot)$ inside it is a small convolutional neural network consisting of two 3x3 convolutional layers, a global average pooling layer, and a final fully connected layer.

\paragraph{4. TGDS}
Eq.~\eqref{eq:tgds} in the main paper shows the topological-aware learning objective of TGDS. The core topological loss $\mathcal{L}{topo}$ is a regularization term based on local consistency, which enforces a topological constraint by penalizing the difference between the pixel prediction $M{pred}(i,j)$ and the average probability of its four neighbors $\mathcal{N}4(i,j)$. The mathematical definition is as follows:
\begin{equation}
    \resizebox{.85\hsize}{!}{$
    \mathcal{L}_{topo} = \frac{1}{HW} \sum_{i,j} \left| M_{pred}(i,j) - \frac{1}{|\mathcal{N}_4(i,j)|} \sum_{\mathbf{p}' \in \mathcal{N}_4(i,j)} M_{pred}(\mathbf{p}') \right|
    $}
\end{equation}

\begin{table*}[htbp]
\tiny
\centering
\caption{Ablation experiment on the sparsity weight $\lambda_{sparse}$.}
\label{tab:supp_sparse_weight}
\resizebox{\textwidth}{!}{
\begin{tabular}{lcccccccccccccccc}
\hline
\multirow{2}{*}{$\lambda_{sparse}$} & \multicolumn{4}{c}{SIRST3-Test} & \multicolumn{4}{c}{NUAA-SIRST-Test} & \multicolumn{4}{c}{NUDT-SIRST-Test} & \multicolumn{4}{c}{IRSTD-1K-Test} \\
\cline{2-17}
& $IoU$ & $nIoU$ & $P_d$ & $F_a$ & $IoU$ & $nIoU$ & $P_d$ & $F_a$ & $IoU$ & $nIoU$ & $P_d$ & $F_a$ & $IoU$ & $nIoU$ & $P_d$ & $F_a$ \\
\hline
0.0010 & 67.18 & 77.65 & 96.45 & 93.81 & 63.13 & 73.69 & 93.61 & 90.15 & 69.19 & 80.59 & 96.98 & 94.98 & 64.73 & 74.25 & 91.89 & 91.31 \\
0.0012 & 67.35 & 77.83 & 96.38 & 93.99 & 63.31 & 73.85 & 93.52 & 90.35 & 69.38 & 80.79 & 96.90 & 95.19 & 64.92 & 74.48 & 91.80 & 91.55 \\
0.0014 & 67.51 & 78.00 & 96.30 & 94.13 & 63.50 & 74.05 & 93.43 & 90.56 & 69.59 & 81.01 & 96.81 & 95.40 & 65.13 & 74.70 & 91.71 & 91.78 \\
0.0016 & 67.69 & 78.18 & 96.22 & 94.28 & 63.71 & 74.24 & 93.34 & 90.75 & 69.78 & 81.20 & 96.73 & 95.59 & 65.34 & 74.91 & 91.62 & 92.00 \\
0.0018 & 67.85 & 78.35 & 96.13 & 94.41 & 63.89 & 74.42 & 93.24 & 90.89 & 69.95 & 81.37 & 96.67 & 95.72 & 65.52 & 75.08 & 91.51 & 92.15 \\
0.0020 & 67.93 & 78.43 & 96.08 & 94.47 & 63.97 & 74.50 & 93.19 & 90.95 & 70.03 & 81.43 & 96.64 & 95.77 & 65.60 & 75.15 & 91.46 & 92.21 \\
\hdashline
0.0040 & 67.98 & 78.47 & 96.06 & 94.50 & 64.01 & 74.54 & 93.17 & 90.98 & 70.06 & 81.46 & 96.63 & 95.79 & 65.63 & 75.18 & 91.44 & 92.23 \\
0.0060 & 68.01 & 78.50 & 96.05 & 94.52 & 64.03 & 74.56 & 93.16 & 91.00 & 70.08 & 81.48 & 96.62 & 95.81 & 65.65 & 75.20 & 91.43 & 92.25 \\
0.0080 & 68.02 & 78.51 & 96.04 & 94.53 & 64.04 & 74.57 & 93.15 & 91.01 & 70.09 & 81.49 & 96.62 & 95.82 & 65.66 & 75.21 & 91.42 & 92.26 \\
0.0100 (\textbf{Ours}) & \textcolor[HTML]{056f28}{\textbf{68.02}} & \textcolor[HTML]{056f28}{\textbf{78.51}} & \textcolor[HTML]{056f28}{\textbf{95.50}} & \textcolor[HTML]{056f28}{\textbf{94.53}} & \textcolor[HTML]{056f28}{\textbf{64.05}} & \textcolor[HTML]{056f28}{\textbf{74.58}} & \textcolor[HTML]{056f28}{\textbf{91.05}} & \textcolor[HTML]{056f28}{\textbf{91.02}} & \textcolor[HTML]{056f28}{\textbf{70.10}} & \textcolor[HTML]{056f28}{\textbf{81.49}} & \textcolor[HTML]{056f28}{\textbf{96.80}} & \textcolor[HTML]{056f28}{\textbf{95.82}} & \textcolor[HTML]{056f28}{\textbf{65.67}} & \textcolor[HTML]{056f28}{\textbf{75.21}} & \textcolor[HTML]{056f28}{\textbf{90.10}} & \textcolor[HTML]{056f28}{\textbf{92.26}} \\
\hdashline
0.0400 & 67.81 & 78.29 & 96.15 & 94.35 & 63.82 & 74.33 & 93.28 & 90.80 & 69.88 & 81.27 & 96.75 & 95.63 & 65.45 & 75.00 & 91.58 & 92.06 \\
0.0800 & 67.43 & 77.91 & 96.34 & 94.05 & 63.41 & 73.95 & 93.48 & 90.43 & 69.49 & 80.89 & 96.86 & 95.28 & 65.05 & 74.61 & 91.75 & 91.69 \\
\hline
\end{tabular}%
}
\end{table*}

\section{Supplementary Details for Experiments}
\label{sec:supp_exp}
This document provides the implementation details and additional ablation experiments omitted for brevity in the main paper \S 4 (Experiment).

\subsection{Implementation Details}
\label{sec:supp_details}
We implement our model based on the PyTorch framework. The backbone of SAM uses the ViT-h version, pre-trained on ImageNet-1K. In Stage One (Knowledge Distillation), we use the AdamW optimizer to train the hierarchical expert mixture adapter for $100$ epochs with a batch size of 32. The initial learning rate is set to $e^{-3}$, and we adopt a cosine annealing strategy. In Stage Two (Knowledge Transfer), we choose SOTA models such as ACM, DNANet, and UIUNet as downstream segmentation networks, also training with the AdamW optimizer for $400$ epochs with a batch size of 16. All experiments are conducted on 8$\times$A100-80G GPUs.

\section{Detailed Component Ablations}
\label{sec:supp_ablations}

\subsection{Expert Combination Effectiveness Analysis}
\label{sec:supp_expert_combo}
We performed detailed ablation experiments on the combination of the four 'white-box' experts to validate the indispensability of each expert. As shown in Table \ref{tab:expert_combination}, we started with the full four-expert model, sequentially removed individual experts, and tested the performance of different expert combinations. The experimental results clearly show that removing any single expert results in a performance decline, and the full four-expert combination achieved the best performance, proving the complementarity and systematic design of our experts.

\subsection{Ablation of Adapter Insertion Layers}
\label{sec:supp_insertion}
The number and placement of Adapter insertions in different layers of SAM's ViT encoder have a significant impact on the experimental results. We conducted ablation experiments on the insertion position and number of Adapters, with results shown in Table \ref{tab:supp_insertion}. The experiments indicate that inserting our MoE Adapter only after the last two ViT blocks achieves the best balance between performance and cost (i.e., "Last 2 ViT Layers (Ours)"). The performance is poorest when the Adapter layers are inserted after all ViT layers, likely due to excessive shallow-layer adapters disrupting SAM's general foundational features.

\subsection{Ablation of Sparsity Loss Weight (\texorpdfstring{$\lambda_{sparse}$}{Lsparse})}
\label{sec:supp_sparse}
We conducted an ablation experiment on the weight coefficient $\lambda_{sparse}$ of the auxiliary sparsity loss. As shown in Table \ref{tab:supp_sparse_weight}, the chosen value of $\lambda_{sparse} = 0.0100$ achieves the optimal performance balance across all datasets.

\end{document}